\documentclass[11pt]{article}
\usepackage{coling2020}
\usepackage{times}
\usepackage{url}
\usepackage{latexsym}
\usepackage{graphicx}
\usepackage{etoolbox}
\usepackage{makecell}
\usepackage{siunitx}
\usepackage{paralist}
\usepackage{amsfonts}
\usepackage{booktabs}
\usepackage{multirow}
\usepackage{subcaption}
\usepackage{tikz}
\usepackage{xspace,mfirstuc,tabulary}
\usepackage{times}
\usepackage{latexsym}
\usepackage{amsmath}
\usepackage{graphicx}
\usepackage{filecontents}
\usepackage{booktabs}
\usepackage{varwidth}
\usepackage{capt-of}
\usepackage{pgfplots,tikz}
\usepgfplotslibrary{colorbrewer}
\pgfplotsset{
	cycle list/Dark2,
	cycle multiindex* list={
		mark list*\nextlist
		Dark2\nextlist
	},
}

\newcommand{\method}[1]{\texttt{#1}\xspace}
\newcommand{\bert}{\method{BERT}}
\newcommand{\berten}{\method{BERT\textsubscript{en}}}
\newcommand{\bertmulti}{\method{BERT\textsubscript{multi}}}
\newcommand{\biobert}{\method{BioBERT}}
\newcommand{\bm}{\method{BM25}}
\newcommand{\tfidf}{\method{TFIDF\textsubscript{char}}}
\newcommand{\wikimesh}{\method{WikiUMLS}}

\newcommand{\recat}[1]{recall@$#1$}
\newcommand{\nrecat}[1]{N. recall@$#1$}
\newcommand{\keq}[1]{$k$$=$$#1$}
\newcommand{\rat}[1]{@$#1$}

\definecolor{bblue}{HTML}{4F81BD}
\definecolor{rred}{HTML}{C0504D}
\definecolor{ggreen}{HTML}{9BBB59}
\definecolor{ppurple}{HTML}{9F4C7C}
\colingfinalcopy 

\title{\wikimesh: Aligning UMLS to Wikipedia via Cross-lingual Neural Ranking}

\author{Afshin Rahimi \\
   School of ITEE\\
  The University of Queensland \\
  Brisbane, Australia \\
  {\tt  a.rahimi@uq.edu.au \enskip} \\\And 
  Timothy Baldwin \quad Karin Verspoor \\
  School of Computing and Information Systems \\
  The University of Melbourne \\
  Melbourne, Australia \\
  {\tt tbaldwin@unimelb.edu.au} \\
  {\tt karin.verspoor@unimelb.edu.au} \\
}

\date{}

\begin{document}
\maketitle
\begin{abstract}
	We present our work on aligning the Unified Medical Language System (UMLS)
	to Wikipedia, to facilitate manual alignment of the two resources. 
	We propose a cross-lingual neural reranking model to match a
	UMLS concept with a Wikipedia page, which achieves a \recat{1} of
	$72\%$, a substantial improvement of $20\%$ over word- and char-level
	BM25, enabling manual alignment with minimal effort. We release our
	resources, including ranked Wikipedia pages for 700k UMLS concepts,
	and \wikimesh, a dataset for training and evaluation of alignment models between UMLS and Wikipedia collected from Wikidata. This will provide easier access to Wikipedia for health professionals, patients, and NLP
	systems, including in multilingual settings.\footnote{Resources available at \url{https://github.com/afshinrahimi/wikiumls}}
\end{abstract}

\section{Introduction}

The Unified Medical Language System (UMLS)\footnote{\url{https://www.nlm.nih.gov/research/umls}}
is a controlled vocabulary
resource, enabling standardisation of biomedical
terminology 
and interoperability of electronic health systems across the
world~\cite{bodenreider2004umls}. UMLS
has good coverage in only a handful languages such as English, impeding its uptake in health
systems in different language settings~\cite{marko2006towards}. In addition, concept definitions in UMLS are either missing or  very short, 
limiting its use as a medical encyclopedia. 
Wikipedia, on the other hand, is a crowd-sourced encyclopedia that is a primary source of
online medical knowledge for practitioners, students, and the general public~\cite{heilman2011wikipedia,heilman2015wikipedia,shafee2017evolution,murray2019more}.
Wikipedia hosts an increasing number of
health-related articles~\cite{matheson2017wikipedia}.
Wikipedia
articles exist in 100s of languages, far exceeding the multilingual support in UMLS. 
Together, these characteristics point to the value of Wikipedia for patient-facing health information.

Our goal in this work is to align UMLS concepts to their corresponding Wikidata and through that to their available Wikipedia concepts, to expand the language support for UMLS terminology with little effort. This will have a direct impact on patients worldwide by facilitating adoption of UMLS (including clinical terminology \method{SNOMED-CT}~\cite{donnelly2006snomed} or Medical Subject Headings, \method{MeSH}~\cite{aronson2010overview}) in international healthcare systems, and also supporting medical information seeking needs of patients with varying linguistic backgrounds. For example, a patient whose native language is not English might receive a discharge summary in English with mentions of symptoms, diagnoses, or medications in formal clinical language. This language may not match Wikipedia titles exactly, making search hard (and noisy). Aligning UMLS and Wikipedia  facilitates such information seeking, makes medical information accessible in a patient's native language, and supports the need for patient access to information in consumer-friendlier terms~\cite{zeng2001patient,smith2005consumer}. 
Most common medical terms are covered in Wikipedia;
for instance, it is estimated that $80\%$ of \method{SNOMED-CT} terms have a corresponding Wikipedia page~\cite{ngo2019can}.
%

Our contributions are as follows: (1) we use the multilingual resources
in UMLS and Wikidata in a neural ranking model to retrieve Wikipedia
articles given a UMLS concept, and achieve a \recat{1} of $72\%$,
representing a $20\%$ increase over a \bm model; (2) we release our new
\wikimesh dataset (collected from Wikidata) for training and evaluation
of UMLS-to-Wikipedia alignment models; and (3) we release the output of
our cross-lingual ranking model as a large-scale silver-standard
alignment of UMLS and Wikipedia, facilitating new applications that can
make use of the alignment.



\section{Background}

UMLS 
combines multiple vocabularies into a unified coding system by mapping
terms referring to the same concept to a single concept ID (CUI). A
single CUI can have several aliases in different languages from various
vocabularies. A large proportion of UMLS concepts (roughly 100k) have descriptions that are 
very short (mostly one sentence), and so are not adequate for dissemination of knowledge or information access.
Medical entity pages in Wikipedia have longer descriptions, are connected to other entities through hyperlinks, and are enriched by the Wikidata~\cite{DBLP:journals/cacm/VrandecicK14} knowledge graph.
Wikipedia is rich in content and multilinguality, 
as anyone
can contribute to or revise an article. For example, the \method{diabetes} Wikipedia page exists in more than $134$ languages, compared to only a dozen languages in UMLS.

\paragraph{Knowledge-base alignment:} There are two main approaches for
aligning knowledge bases such as Wikipedia and UMLS: (1) embedding-based
alignment, and (2) string and semantic matching. In embedding-based
methods, entity embeddings are learnt from text co-occurrence statistics
or knowledge-graph (KG) relations separately, and are then aligned using
a seed alignment
dictionary~\cite{mikolov2013exploiting,DBLP:conf/ijcai/ChenTYZ17}, or
adversarial learning~\cite{qu2019weakly}. However, the text-based
methods suffer from non-comparable
corpora~\cite{ormazabal-etal-2019-analyzing}, and the KG methods have
not been tried on Wikidata because of its large scale and entity
variety.

String and semantic matching methods are based on
similarity between the entity names or descriptions in the two knowledge-bases (KBs). Each
entity in $\mathrm{KB}_1$ is used as a query against all the entities in
$\mathrm{KB}_2$ using a scoring method such as \method{cosine} similarity or \method{BM25}. Entities are represented by bag of words or bag of character $n$-grams.  
The result of these cheap methods can be further improved by supervised (neural) ranking models~\cite{DBLP:conf/cikm/GuoFAC16,rao2019bridging,DBLP:conf/bionlp/WangBNLWA18}. Particularly
when query and candidate documents are represented by an encoder such as
BERT~\cite{devlin-etal-2019-bert}, pre-trained on massive
amounts of text data, neural rerankers perform substantially
better than IR-based methods~\cite{DBLP:journals/corr/abs-1901-04085,akkalyoncu-yilmaz-etal-2019-cross}. Our methods fit within this group. 

\section{Method}

Given a UMLS concept $c_i$ represented by query
$q_i=\{t^1_i,\dots, t^N_i\}$, where $t^n_i$ is an alias term for $c_i$
in UMLS, we use the English Wikipedia augmented with multilingual Wikidata
as a document collection $D=\{d_1,\dots,d_{\vert D \vert} \}$, to 
retrieve page $d_j$ matching concept $c_i$. Each page is represented
by its title, text (only for candidate generation), and multilingual aliases from Wikidata. 
 We follow a two-stage retrieval procedure: (1) candidate generation, where
an IR method (e.g.\ \bm) is used to retrieve related documents; and
(2) reranking of the top $k$ candidates via a learn-to-rank method~\cite{liu2009learning}.

\subsection{Candidate Generation}

We index Wikipedia collection $D$ using \method{Lucene}, and build query
$q_i$ from UMLS to retrieve the top \keq{64} relevant pages. We use a
Boolean disjunction between all alias terms in UMLS, and search in the
title, text, and multilingual aliases fields in $D$.  \bm relies on
exact term matches, and small variations can result in a mismatch. As a
result, we also experimented with a character $n$-gram method (\tfidf)
successfully used in \newcite{murty-etal-2018-hierarchical} for candidate
generation in medical entity linking. We build a bag of character $n$-grams
($n \in [1, 5])$) weighted by \method{TF-IDF} within term boundaries,
and use cosine similarity between $q_i$ and each $d \in D$ (excluding
page text) to generate the top \keq{64} candidates.

\subsection{Reranking} 

We formulate the reranking task as passage pair binary
classification~\cite{DBLP:journals/corr/abs-1901-04085}, where the first
passage is $q_i$ for concept $c_i$ from UMLS, and the second passage is
the set of Wikipedia alias names for each of top 
\keq{64} documents ranked by candidate
generation (Wikipedia text is excluded for reranking). The goal is to predict if a pair is a match or not, by
minimising the following objective:

\vspace{-0.6cm}
\begin{align*}
L(q_i, \mathrm{cand}_i) =  - \log (f(q_i, d^+)) 
- \sum_{d^- \in \mathrm{cand_i}^-} \log (1 - f(q_i, d^-))
\end{align*}
\vspace{-0.4cm}

\noindent where $(q_i, d^+)$ is the matching UMLS-Wiki pair, and
$\mathrm{cand}_i^-$ is the set of remaining negative candidates
generated by \bm. Function $f$ is the passage pair encoder, for which we
use \bert's \method{<CLS>} token
encoding~\cite{devlin-etal-2019-bert} and a linear projection. We also experiment with
\biobert~\cite{10.1093/bioinformatics/btz682} because it is
pre-trained on medical literature, which is a better domain fit for our
task.

The major shortcoming of \bert and \biobert is that they don't encode
multilingual aliases effectively, particularly if the scripts are
different. 
Aliases of every concept in UMLS and Wikipedia are available on average in $1.4$ and $11.2$ different languages, respectively. This multilingual data can be utilised for triangulation of the matching pair. For example, \textit{GERD} is a term used in UMLS for Gastroesophageal reflux disease, and is also a given name in Germanic languages. If it is used for retrieval, many Wikipedia pages related to people with that name will rank high. However, if alias names of \textit{GERD} in other languages (e.g.\ Japanese) are used in the query and documents, the disambiguation becomes easier. To encode multilingual alias concept names we require a model that embeds tokens of different languages into the same embedding space. To this end, we also experiment with multilingual BERT (\bertmulti).

\begin{table}[t!]
	\centering
	\begin{tabular}{l p{0.35\linewidth} p{0.35\linewidth} }
		\toprule
		UMLS Concept & UMLS Query & Gold Wiki Candidate \\
		\midrule	
		CUI: C0017168 & GERD, Acid Reflux, oesofagusaandoening, \dots & disorder of the esophagus, enfermedad esofagal, \dots \\
		\bottomrule
	\end{tabular}
	\caption{The multilingual UMLS aliases are used as a query against all Wikipedia entities to generate $k$ candidates. The candidates are then reranked using a neural model.}
	\label{tab:example}
\end{table}



\subsection{Data}
\textbf{\wikimesh} is a UMLS to Wikipedia aligned dataset we create in this work. It consists of about 17.8k UMLS concepts that are manually linked to their matching Wikipedia page by Wikipedia content contributors. All the UMLS concepts in \wikimesh are from Medical Subject Headings vocabulary (MeSH)~\cite{lipscomb2000medical}. A \wikimesh record is a tuple of (\method{UMLS CUI}, \method{UMLS concept alias set}, \method{Wikipedia page title}, \method{Wikidata concept alias set}). We use a UMLS concept's alias set as query, and compare it with every Wikidata entity's alias set to retrieve the matching Wikipedia page. The Wikipedia aliases and their links to UMLS are taken from their corresponding record in Wikidata, a collaborative knowledge-base that is tightly connected to Wikipedia. An entity in Wikidata has multilingual aliases,\footnote{For the purposes of this work, we focus on the 24 languages currently supported in UMLS: en, es, it, nl, fr, pt, de, cs,
	ru, zh, ja, hu, tr, ko, no, et, sv, pl, fi, el, lv, da, eu, he.} and is linked to both a Wikipedia page, and possibly a UMLS concept.\footnote{A sample Wikidata record:
	\url{https://wikidata.org/wiki/Q12206}.}
We split the matching pairs into roughly 10k, 2k, and 5.8k for the training,
validation, and test data sets, respectively. There are about 3 million
remaining UMLS concepts from various vocabularies that are not aligned
to Wikipedia, which we hope to align in this work. 




\subsection{Evaluation Methodology}

We use the aliases of a concept in UMLS as a query, and the aliases for
entities in Wikipedia as the document collection. A document is relevant to a query if the pair are manually aligned in \wikimesh (a matching UMLS-Wikipedia pair). We evaluate the models by measuring recall at different positions of their ranking (\recat{k}). In this paper, recall and accuracy are equivalent because we assume each UMLS concept has a unique matching Wikipedia page. 
Because our proposed retrieval method has two stages (candidate generation followed by neural reranking), we use normalised recall (\nrecat{k}) to evaluate the neural reranking stage independent of candidate generation stage by ignoring queries for which the candidate generator didn't retrieve a relevant document.

\begin{figure}[t!]
  \begin{minipage}{\textwidth}
	\begin{minipage}[b]{0.40\textwidth}
		\centering
		\begin{tikzpicture}[trim axis left]
	\begin{axis}[
	width=1.2\textwidth,
	grid=both,
	grid style={dashed,gray!30},
	axis y line*=left,
	axis x line*=left,
	ylabel near ticks,
	xlabel near ticks,
	xlabel=$k$,
	ylabel=\recat{k},
	legend pos=south east,
	legend cell align={left},
	xtick={1, 2, 4, 8, 16, 32, 64},
	xticklabels={1, 2, 4, 8, 16, 32, 64},
	xmode=log,
	log basis x=2,
	]
	
	\addplot coordinates {
		(1, 0.67) (2, 0.74) (4, 0.79) (8, 0.82) (16, 0.84) (32, 0.85) (64, 0.85)
	};
	
	\addplot coordinates {
		(1, 0.72) (2, 0.78) (4, 0.81) (8, 0.83) (16, 0.84) (32, 0.85) (64, 0.85)
	};
	
	\addplot coordinates {
		(1, 0.68) (2, 0.76) (4, 0.80) (8, 0.83) (16, 0.84) (32, 0.85) (64, 0.85)
	};

	\addplot coordinates {
		(1, 0.59) (2, 0.67) (4, 0.73) (8, 0.77) (16, 0.80) (32, 0.83) (64, 0.85)
	};

	\addplot coordinates {
		(1, 0.56) (2, 0.63) (4, 0.67) (8, 0.71) (16, 0.74) (32, 0.76) (64, 0.79)
	};
	\legend{\berten, \bertmulti, \biobert, \bm, \tfidf}
	\end{axis}
	\end{tikzpicture}
	\label{fig:results}
		\captionof{figure}{\recat{k} for neural reranking models (\bertmulti, \berten, \biobert),
			and word- (\bm) and character-level (\tfidf) models.} 
	\end{minipage}
	\hfill
	\begin{minipage}[b]{0.57\textwidth}
		\centering
		\begin{tabular}{lcccccc}
		& \multicolumn{3}{c}{Recall} && \multicolumn{2}{c}{N. Recall} \\
		\cline{2-4}  \cline{6-7}
		Method & \rat{64} & \rat{4} & \rat{1} & & \rat{4} &  \rat{1}\\
		\midrule
		\bm & 85 & 73 & 59 && --- & ---\\
		\tfidf & 79 & 67 & 56 && --- & ---\\
		\midrule
		\berten & 85 & 79 & 67 && 92 & 78\\
		\biobert  & 85 & 80 & 68 && 94 & 79\\
		\bertmulti & 85 & \textbf{81} & \textbf{72} && \textbf{95} & \textbf{84}\\
		\label{tab:results}
	\end{tabular}
		\captionof{table}{Performance of the vector space models (\bm, \tfidf),
			and neural ranking models (\berten, \biobert, \bertmulti) over
			Wikipedia page retrieval given a UMLS concept. 
			Normalised recall (\nrecat{k}) is calculated by excluding queries
			for which the candidate generator (\bm) didn't retrieve the
			relevant document.}
	\end{minipage}
\end{minipage}
  
\end{figure}

\section{Results}

The performance of the candidate generation methods for word- (\bm) and
character-level (\tfidf) retrieval, and also the reranking 
methods (\bert,
\biobert, and \bertmulti) are shown in Figure 1. In all top
$k$ positions, \bm outperforms \tfidf, which is surprising given
that \tfidf was reported to achieve strong performance for entity linking in
\newcite{murty-etal-2018-hierarchical}.\footnote{The \recat{100} of
	\tfidf for entity linking candidate generation in
	\newcite{murty-etal-2018-hierarchical} is $61.09\%$, much higher than
	the results for our re-implementation over the same dataset, possibly due to differences in implementation or the candidate set.}  
As shown in
Table 1, \bm is able to retrieve the correct Wikipedia page
at \keq{64} with $85\%$ accuracy (compared to $75\%$ for \tfidf), an
upper-bound for the performance of the reranking models.

For the reranking models, \biobert performs only slightly better than
\bert ($68\%$ vs.\ $67\%$ \recat{1}), although it has been pretrained on large amounts of biomedical
literature. \bertmulti performs better than both \bert and
\biobert, achieving a \recat{1} of $72\%$, given the cross-lingual nature
of the task.  Compared to classic vector space approaches (e.g.\
\bm), \bertmulti shows an improvement of $22\%$, and $10\%$ for
\recat{1} and \recat{10}, respectively. We also report normalised recall at
\keq{1} and \keq{4}, by excluding the test instances for which \bm doesn't
retrieve the gold candidate. Here, \bertmulti achieves normalised recall at
\keq{1} and \keq{4} of $84\%$ and $95\%$, respectively. This indicates
that \bertmulti is highly successful at ranking in the case that the
correct document is retrieved.



\section{Conclusions}

We proposed passage pair ranking models based on pretrained contextual
encodings for aligning UMLS and Wikipedia, to help bridge between health information systems, and
empower consumers with understanding of their health condition. We developed a dataset,
\wikimesh, for training and testing alignment models between the two
knowledge-bases, and proposed neural reranking models that 
substantially outperform \bm. We showed that the use of
multilingual aliases in \bertmulti substantially improves \recat{1}
compared to \biobert ($72$ vs.\ $68$).

The use of subword information such as
\method{BPE}~\cite{sennrich-etal-2016-neural} as used in
\method{XLM}~\cite{conneau2019cross} might improve performance, which we
leave for future work. Utilising the relationships between concepts in UMLS and Wikipedia (through Wikidata) to align the two knowledge graphs is also an interesting future direction. We also intend to release a large Wikipedia-based Entity
Linking (EL) dataset by using the top-ranked Wikipedia pages for UMLS queries, to be used in state-of-the-art EL models such as
\method{zeshel}~\cite{logeswaran-etal-2019-zero}.

\section*{Acknowledgements}
This work was funded by the Australian Research Council through the ARC
Training Centre in Cognitive Computing for Medical Technologies (project
number ICI70200030)), and completed while the first author was a post-doctoral researcher at The University of Melbourne.
\bibliographystyle{coling}
\bibliography{coling2020}

\end{document}